%% file: main.tex
\documentclass[10pt]{article}

\newif\ifarxiv

\arxivtrue

\ifarxiv 
\twocolumn
\date{}

\def\ninept{\def\baselinestretch{.95}\let\normalsize\small\normalsize}

\else 
\usepackage{spconf}
\fi

\usepackage{amsmath,graphicx}

\usepackage{xspace}
\usepackage{booktabs}
\usepackage{color}
\usepackage[dvipsnames]{xcolor}
\usepackage{setspace,placeins}
\usepackage{colortbl}
    
\usepackage{amsfonts}
\usepackage{graphicx}
\usepackage{pifont}
\usepackage{multirow}
\usepackage[letterspace=10]{microtype}
\usepackage{palatino} 

\usepackage{subcaption}

\newcommand{\ie}{\emph{i.e.}\@\xspace}
\newcommand{\etal}{\emph{et al}\@\xspace}

\newcommand{\kenza}[1]{{\color{magenta}[\textbf{Kenza}: #1]}}

\newcommand{\mypar}[1]{\bigskip \medskip \noindent \textbf{#1} \hspace{1.5em}}

\def \etal    {\textit{et al.}\xspace}

\def \Tset    {{\mathcal T}}
\def \Mset    {{\mathcal M}}

\def \Cset    {{\mathcal C}}

\def \Kset    {{\mathcal K}}
\def \Qset    {{\mathcal Q}}

\title{Nearest neighbor search with compact codes: A decoder perspective}
\ifarxiv
\author{Kenza Amara,
Matthijs Douze, 
Alexandre Sablayrolles,
Herv\'e J\'egou}

\else
\name{
Kenza Amara,
Matthijs Douze, 
Alexandre Sablayrolles,
Herv\'e J\'egou
}
\address{Facebook AI}
\fi
\begin{document}

\ninept
\renewcommand{\arraystretch}{1.12}
\renewcommand{\baselinestretch}{1.2}

\maketitle
\begin{abstract} 
Modern approaches for fast retrieval of similar vectors on billion-scaled datasets rely on compressed-domain approaches such as binary sketches or product quantization. 
These methods minimize a certain loss, typically the mean squared error or other objective functions tailored to the retrieval problem. 
In this paper, we re-interpret popular methods such as binary hashing or product quantizers as auto-encoders, and point out that they implicitly make suboptimal assumptions on the form of the decoder. 
We design backward-compatible decoders that improve the reconstruction of the vectors from the same codes, which translates to a better performance in nearest neighbor search. 
Our method significantly improves over binary hashing methods or product quantization on popular benchmarks. 
\end{abstract}
\ifarxiv 
\else 
\fi

\input{intro.tex}

\input{related.tex}

\input{method.tex}

\input{experiments.tex}

\section{Conclusion}

In this paper we have focused on the decoder associated with popular approximate nearest neighbor search based on compact codes. 
Our main proposal is to design stronger decoders for existing encoders for approximate search. 
We have evidenced that decoders associated with existing methods are suboptimal in terms of reconstruction given the indices. 
We have proposed an enhanced decoder based on a neural network that we use with several types of encodings, such as binary hashing method or product quantization. This optimized decoder improves the accuracy when performing similarity search, and we do not compromise the efficiency since the main use-case of our method is to provide a re-ranking stage. 

\bibliographystyle{abbrv}

\bibliography{bibliography}

\renewcommand{\arraystretch}{0.92}
\renewcommand{\baselinestretch}{0.92}
%

\label{sec:refs}

\end{document}

%% file: intro.tex
\section{Introduction}
\label{sec:intro}

The emergence of large-scale databases raise new challenges, one of the most prominent ones being on how to explore efficiently this data. 
Finding similar vectors in  large sets is increasingly important with the emergence of vector \emph{embeddings} that represent data of various modalities~\cite{douze20212021,mikolov2013efficient}. 
Exact nearest-neighbor search in high-dimensional spaces is intractable~\cite{weber1997approximation}, which is why researchers and practitioners have resorted to approximate nearest-neighbors (ANN), trading some search accuracy against orders of magnitude gains in response time, and memory consumption. 
Amongst the techniques widely adopted in industry~\cite{johnson2019billion,guo2020accelerating,subramanya2019diskann}, quantization-based approaches~\cite{matsui2018survey}, like product~\cite{jegou2010product} or additive quantizers~\cite{babenko2014additive,liu2015improved,martinez2016revisiting,martinez2018lsq++}, estimate distances based on approximated vector representations.

In this paper, we regard search methods based on compact codes as auto-encoders, and address the problem of improving the decoder for a fixed encoder: we assume that the stage that assigns vectors to codes is fixed, and we examine how to improve decoding if we tolerate some runtime impact. 
This setting is especially useful in situations where (1) we need backward-compatibility on existing codes, and/or (2) for re-ranking to refine an initial short-list~\cite{jegou2011searching}. 

The motivation behind our method is to exploit the inherent suboptimality of existing decoders, which typically assume that there is no residual mutual information between bits or subindices. 
Lifting this assumption, we design a decoder that offers a better estimation of the reproduction value (or centroid) associated with binary sketches or structured compact codes employed in multi-codebook quantization. 
We demonstrate the potential of uncoupling the encoder and decoder for several effective encoders such as binary codes~\cite{gong2012iterative,sablayrolles2018spreading} or product quantization~\cite{jegou2010product}. 
Our solution relies on a simple neural decoding network. On the BigANN~\cite{jegou2011searching} and Deep1M~\cite{babenko2016efficient} benchmarks, it provides substantial gains w.r.t. the trade-off between reconstruction and memory budget. 
Noticeably, we use a very efficient encoder for index construction and initial search, like a binary or fast quantizer~\cite{andre2019quicker}, and use our neural decoder to re-rank a short-list with high-quality neighbors.

%% file: related.tex
\section{Preliminaries}
\label{sec:related}

In this section, we first present the quantization methods involved approximate nearest neighbor search as auto-encoders. We then discuss popular quantization methods for which our paper proposes to improve the decoder while keeping the encoder fixed. 

\subsection{Quantization techniques for ANN search}

Most vector encoding methods for approximate nearest neighbor search can be interpreted as quantization techniques~\cite{gray1998quantization}. 
A quantizer can be regarded as an auto-encoder of the form 
\begin{align} \label{eq1}
    x \xrightarrow{\normalsize\ \ f\ \ } 
    f(x)=k \in \Kset  \xrightarrow{\ \ g\ \ } 
    q_k = g(f(x)) \in \Qset \subset \mathbb R^d,  
\end{align}
where the input vector $x \in {\mathbb R}^d$ is first mapped by an encoder $f$ into a code $k \in \Kset$.
The encoder $f$ implicitly defines a partitioning of $\mathbb{R}^d$ into $K=|\Kset|$ disjoint cells $\Cset_1, \dots, \Cset_K$, where $\Cset_k = f^{-1}(k)$. 
The decoder reconstructs an approximation $q_k$ from the code $k$, which belongs to the set $\Qset=\{q_k\}_{k \in \Kset}$ of reproduction values. 
The encoder-decoder $q=g \circ f $ is usually referred to as a quantizer~\cite{gray1998quantization}. 

\mypar{Lloyd's optimality conditions.}
Given cells and their corresponding reproduction values $q_k$, Lloyd~\cite{lloyd1982least} derived two necessary conditions for a quantizer to be optimal in terms of the average squared loss. 
First $x$ must be assigned its closest reproduction value, 
which translates to the usual assignment rule to the nearest centroid: 
\begin{equation}
    f(x) = \mathrm{argmin}_{k' \in \Kset}~\| x - q_{k'} \|^2. 
\end{equation}

This condition defines the optimal quantizer for a given set of reproduction values, whether we can enumerate it or not. 
Denoting by $p$ the p.d.f. of the input data, the second condition is that each reproduction value $q_k$ should be the expectation of the vectors assigned to the same cell as 
    \begin{equation}
    q_k = \int_{x \in \mathbb \Cset_k}  p(x) x  dx, 
    \label{equ:optimal_reconstruction}
    \end{equation}

\subsection{Structured vector quantization }
The most general form of vector quantization is when the set of reproduction values $\Qset = \{q_1, \dots, q_K\}$ is unconstrained, such as the one typically produced by  k-means. 
In the context of coding for distance estimation, a very large number of centroids (typically, $2^{128}$) is required to obtain a sufficient precision.  
It is not feasible to run k-means at that scale.

\mypar{Product Quantization (PQ). }
In order to learn fine-grained codebooks, J\'egou et al.~\cite{jegou2010product,sandhawalia2010searching} propose a product quantizer, where the set of centroids $\Qset$ is implicitly defined as a Cartesian product of $m$ codebooks $\Qset = \Qset_1 \times ... \times \Qset_m$. 
Each codebook $\Qset_i$ consists of $K'$ centroids defined in $\mathbb R^\frac{d}{m}$. 
The assignment is separable over the $m$ subspaces and produces indexes of the form $k=(k_1,\dots,k_m)$. 
The advantage is that the total number of centroids is $(K')^m$ with an assignment step to centroid with an efficient complexity in ${\mathcal O}(dK')={\mathcal O}(dK^\frac{1}{m})$, where $d$ denotes the vector dimensionality.

\mypar{Notation PQm$\times$b.} 
We denote by PQm$\times$b a product quantizer defined by $m$ subquantizers with b-bits subindices. It corresponds to a compact code of size m$\times$b.

\mypar{Additive quantizers (AQ)} generalize this, they define the reproduction values as 
\begin{equation}
    \Qset = \{ c_1 + \dots + c_m | c_1 \in \Qset_1, \dots, c_m \in \Qset_m \}, 
    \label{equ:additive_quantizer}
\end{equation}
where $\forall i\ \Qset_i \subset \mathbb R^d$. 
Similar to product quantization, the indices are tuples. 
When not ambiguous, we use notation $\Qset_i[k]$ for the element indexed by $k$ in $\Qset_i$. 
Functions implemented as look-up tables (LUTs) can be written as: 
\begin{equation}
    x \xrightarrow{\normalsize\ \ f\ \ } 
        \begin{pmatrix}
    k_1=f_1(x) \\
    \vdots \\
    k_m = f_m(x) \\
    \end{pmatrix}
    \xrightarrow{\ \ g\ \ } 
    g(f(x)) = \sum_{i=1}^m \Qset_i[k_i].  
    \label{equ:additive_quantizer2}
\end{equation}

There are different forms of additive quantizers, with different encoder algorithms: the form of their decoders is identical and rely on LUTs as in Eqn.~\ref{equ:additive_quantizer}.  
For instance for a residual quantizer~\cite{liu2015improved} the assignment is done sequentially, which is fast but does not guarantee to assign a vector to its closest neighbors. 
Subsequent additive quantizers, like the ones by Babenko et al.~\cite{babenko2014additive}, and Local Search Quantization (LSQ)~\cite{martinez2016revisiting,martinez2018lsq++} by Martinez et al. improve the trade-off between encoding complexity and reconstruction error. 

\mypar{Optimal centroids for a fixed encoder.}  
Given a set of training vectors $(x_i)_{i=1..n}\in\mathbb{R}^d$ and their codes $k_1,...,k_m$, it is possible to construct an additive decoder (Eqn.~\ref{equ:additive_quantizer}) that minimizes the $\ell_2$ loss.
Denoting by $X\in \mathbb{R}^{n\times d}$ the matrix of training vectors, $C\in \mathbb{R}^{mK'\times d}$ the codebook entries, and converting subindices $k_1,...,k_m$ into one-hot vectors stacked in $I \in \{0, 1\}^{n\times mK'}$, the optimal solution~\cite{babenko2014additive,martinez2018lsq++} is given by
\begin{equation}
\mathrm{argmin}_C \|X - CI\|_{2}^{2} + \lambda \| C \|^2 _2, 
\label{equ:aq}
\end{equation}
where the first term minimizes the reconstruction error on the training set. 
As noted by Martinez \etal~\cite{martinez2018lsq++}, this estimation has numerical stability issues, which is addressed with the regularizer weighted by $\lambda > 0$. %
This minimization is performed component-wise~\cite{babenko2014additive} in closed form and is therefore efficient to obtain.

\mypar{Distance estimator.} 
At search time, the ANN algorithm estimates the distance $d(x,y)$ or similarity between a query $x$ and each database vector $y$ based on an imperfect representation of $y$ or both $x$ and $y$. 
When both the query and database vectors are quantized, it is a Symmetric Distance Comparison (SDC), which approximates any square distance $d(x,y)^2$ by the estimator
\begin{equation}
d_\textrm{SDC}(x,y)= d(q(x),q(y))^2. 
\end{equation}
The asymmetric distance computation (ADC)~\cite{jegou2010product} estimates distances as 
\begin{equation}
d_\textrm{ADC}(x,y)=  d(x,q(y))^2. 
\end{equation}
In this case the query vector $x$ is not quantized. 

Note that the quantization is a lossy operation: the quality of neighbors strongly depends on the estimator and of the quantizer. 
ADC reduces the quantization noise compared to SDC, which subsequently improves the search quality~\cite{jegou2010product}.

\mypar{Compressed-domain calculation.} 
For both PQ and AQ, the comparison is performed in the compressed domain, one does not need to decompress the database vectors explicitly as discussed by J\'egou et al.~\cite{jegou2010product} and~\cite{babenko2014additive}.

\subsection{Hashing based ANN}
Binary codes are quantization techniques that 
derive from Locality-Sensitive hashing (LSH) \cite{charikar2002similarity,indyk1998approximate, gionis1999similarity}. 
In this work we focus on binarization, which ensures that a small Hamming distance between bit vectors implies proximity in the original space for a given metric, for instance cosine~\cite{charikar2002similarity}. 
Binarization maps a vector $x$ to a sequence of bits $(k_1,\dots,k_m)$ using $m$ elementary projections $u_i$: $k_i=\textrm{sign}(u_i^\top x)$. 
It is a form of quantization where the reconstruction is possible up to some scaling constant. 
If the $\{u_i\}_i$ is an orthonormal set, then the reconstruction on the unit-norm $\ell_2$-hyper-sphere as  %
\begin{equation}
    q_k = \frac{1}{\sqrt d} \sum_{i=1}^m  k_i u_i \propto [u_1,\dots,u_m] \begin{bmatrix} k_1 \\[-5pt] {\scriptsize \vdots} \\[-2pt] k_m \end{bmatrix}. 
    \label{equ:recons_binary}
\end{equation}
leads to the same ranking as the Hamming distance between the binary $k_i$. Note, we use an explicit reconstruction to compute ADC for binary vectors. 

We consider two training methods for ANN search with binary codes. The first is Iterative Quantization~\cite{gong2012iterative} (ITQ). This simple embedding (learned rotation and sign selection) serves as a baseline in numerous publications. The second is the catalyzer of Sablayrolles \etal~\cite{sablayrolles2018spreading}, which produces high-quality binary embeddings with a neural network.  
We refer the reader to existing reviews for other approaches~\cite{wang2014hashing,wang2015learning}.

\subsection{Re-ranking methods}
Some Locality-Sensitive Hashing algorithms such as E2LSH~\cite{datar2004locality} rely on a two-stage approach, where (1) a first system selects the most promising neighbor candidate; (2) which are filtered out by a  re-ranking system exact distance computation. 
The VA-file~\cite{weber1997approximation} is  the ancestor of approximation-based filtering: a first approximation of the vector leads to select a short-list of neighbor candidates. This approximation being too crude, a re-ranking stage computes the exact distance between the query and the exact representation of the vectors in the short-list. 
This involves a significant amount of extra storage for large databases. Some approaches alleviate this constraint by refining the first-stage approximation with a secondary compact code~\cite{jegou2011searching}. 

\subsection{Architectural considerations} 
Indexing algorithms heavily depend on the hardware on which they are run. 
Compared to other quantization approaches based on compact codes, binary hashing is less precise but  benefits from specific low-level instructions of modern CPUs, like \texttt{XOR} and \texttt{popcount} that make the distance computation very fast. 
Quantization methods significantly benefit from algorithms running on the GPU~\cite{johnson2019billion}. With smaller PQ codebooks, order(s) of magnitude faster distance comparisons can be obtained by computing ADC distance in registers~\cite{andre2019quicker}. 
This requires to adopt smaller quantization codebooks. 
For instance, K'=16 instead of the more standard setting K'=256 with product or additive quantization.

%% file: method.tex
\section{Method}
\label{sec:method}

Our proposal improves the decoder given an existing encoder, such that our decoder can be used in a re-ranking stage to improve the ranking. There are several advantages to keep a fast encoder in approximate search techniques based on compact codes, noticeably a faster indexing and large-scale search. %

\begin{figure*}%
    \centering
    \includegraphics[width=0.6\linewidth,clip,trim=0 0pt 0pt 12pt]{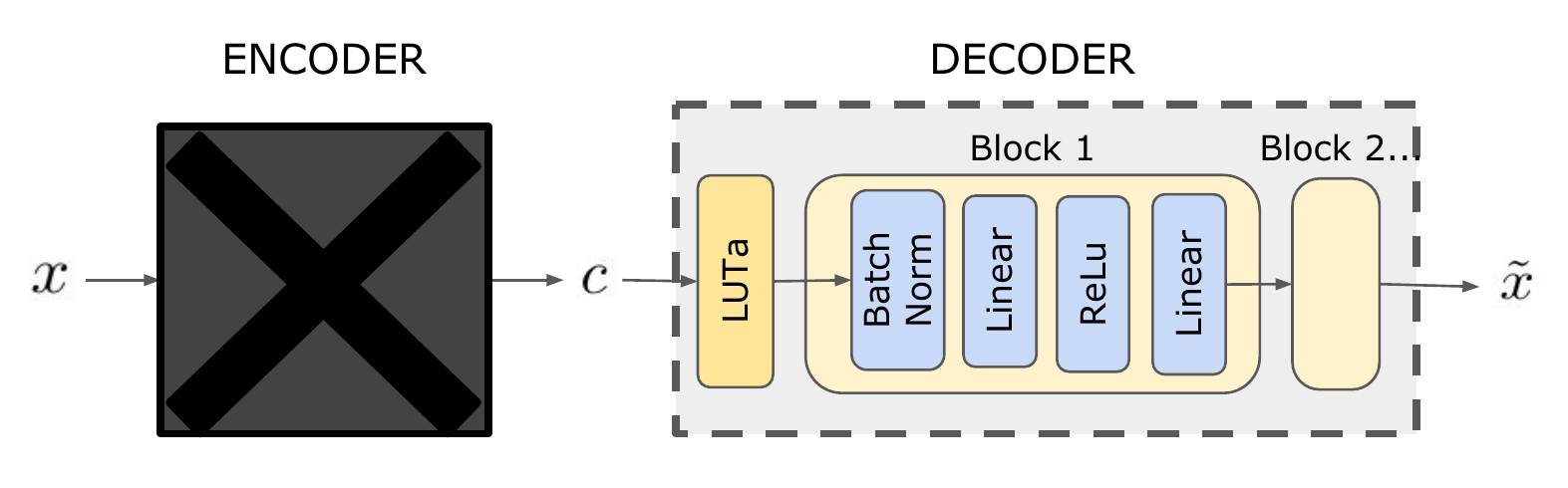}
    \caption{Neural net decoder architecture. The encoder is fixed and we train the decoder to minimize the loss $\|\tilde{x}-x\|^2$ and/or a triplet loss. The first layer of the decoder has the structure of an additive look-up table (LUTa).
        \label{fig:nn_linear}}
\end{figure*}

In this section we first introduce the binary and quantization-based encoders that we focus on.  We evidence sub-optimalities in existing approaches on a simple case with a tractable optimal decoder. 
Then we introduce our approach based on a neural network decoder (denoted NN) illustrated in Figure~\ref{fig:nn_linear}, which we adopt with any type of encoder. 

\subsection{Towards stronger decoders \& a discussion} 

In Table~\ref{tab:alternative_Decoders} we give the set of encoders that we consider: we consider popular and state-of-the-art binarization and quantization methods. We indicate the usual decoder and provide their standard decoder in the column ``decoder'' along with our replacement proposal in the column ``proposed decoder''. 

For Product Quantization (denoted PQ) and binary codes, we consider 64 bits codes for a more direct comparison with the literature. 
We denote by PQ8$\times$8 the usual product quantizer defined by $m$\,=\,$8$ subquantizers with 8-bits subindices (i.e., $K'$\,=\,$256$) and by PQ16$\times$4 a product quantizer such that $m$\,=\,$16$ and $K'$\,=\,$16$. 

\begin{table}[t]
\caption{Encoder-decoder considered in this paper, and proposed decoders that we propose instead for ranking or re-ranking. %
Our proposal is to design a stronger decoder for each binarization or quantization technique: either we compute the optimal look-up tables (w.r.t. reconstruction), as initially proposed for additive quantization (AQ), or we train a neural network decoder (NN). %
\label{tab:alternative_Decoders}}
\centering {  %
\begin{tabular}{lcc}
\toprule
encoder & decoder & proposed decoders \\
\midrule
ITQ       ~\cite{gong2012iterative}                 & naive  & AQ, NN  \\
Catalyzer ~\cite{sablayrolles2018spreading}& naive  & AQ, NN   \\
PQ16$\times$4    ~\cite{jegou2010product,andre2019quicker} & PQ     & AQ, NN   \\ 
PQ8$\times$8     ~\cite{jegou2010product}                  & PQ     & AQ, NN  \\ 
LSQ++~\cite{martinez2018lsq++}                      & AQ     & --  \\
\bottomrule
\end{tabular}}
\end{table}

\begin{figure}[t]
\includegraphics[width=\linewidth,trim=0pt 0pt 25pt 15pt, clip]{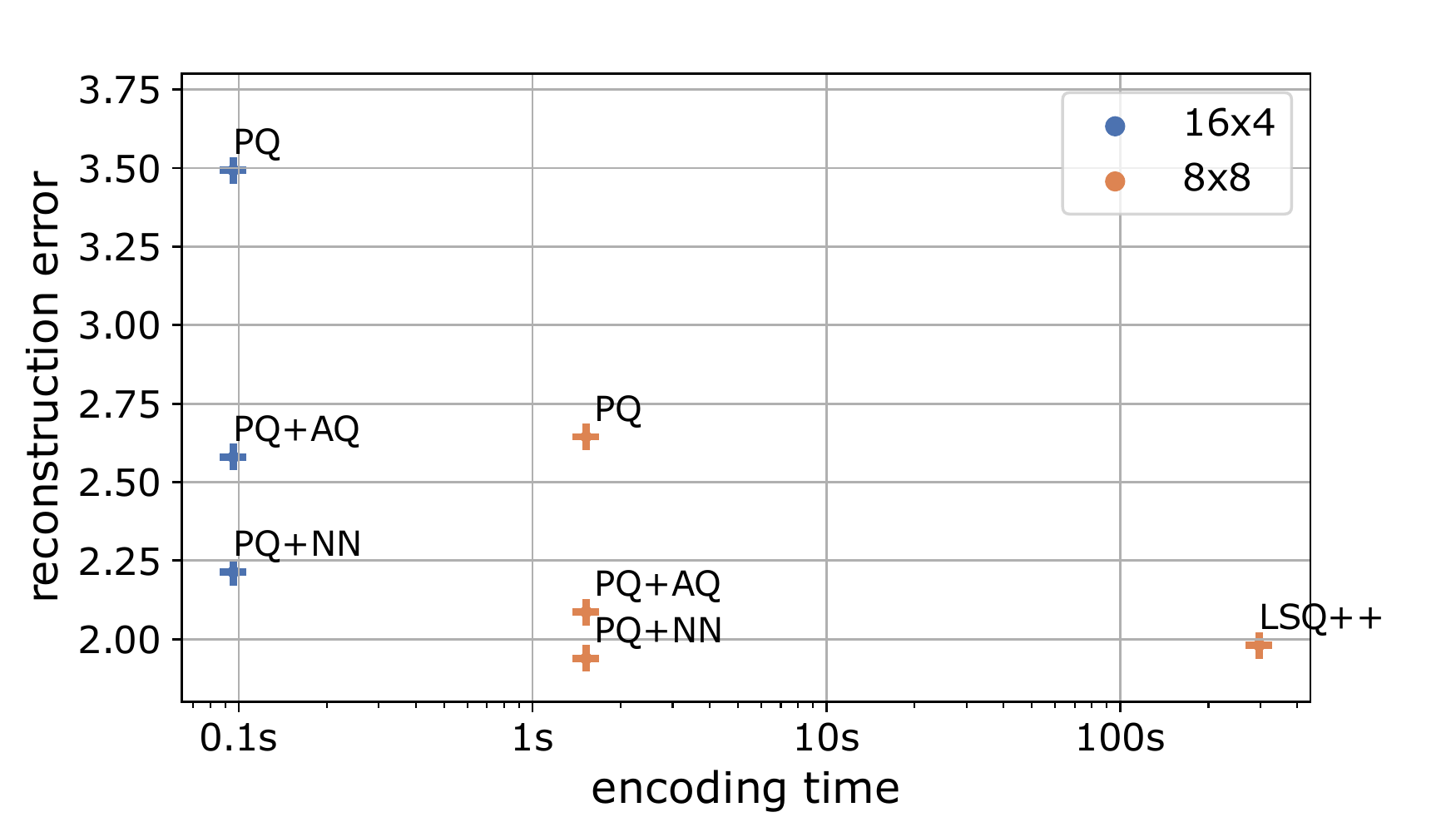}%
\vspace{-0.5em}
\caption{Trade-off between MSE and encoding time for multiple quantizers. LSQ++ (8$\times$8) outperforms regular PQ w.r.t. MSE, but the encoding speed is prohibitively slow. In contrast, PQ16$\times$4  allows for a quick encoding but has a poor reconstruction. We improve the compromises by changing the decoder for a given encoder (PQ+AQ \& PQ+NN). 
}
\label{fig:encoding_time}
\end{figure}

\begin{figure*}[t]
\hspace{1cm} \includegraphics[height=0.27\linewidth,trim=15pt 0 55pt 10, clip]{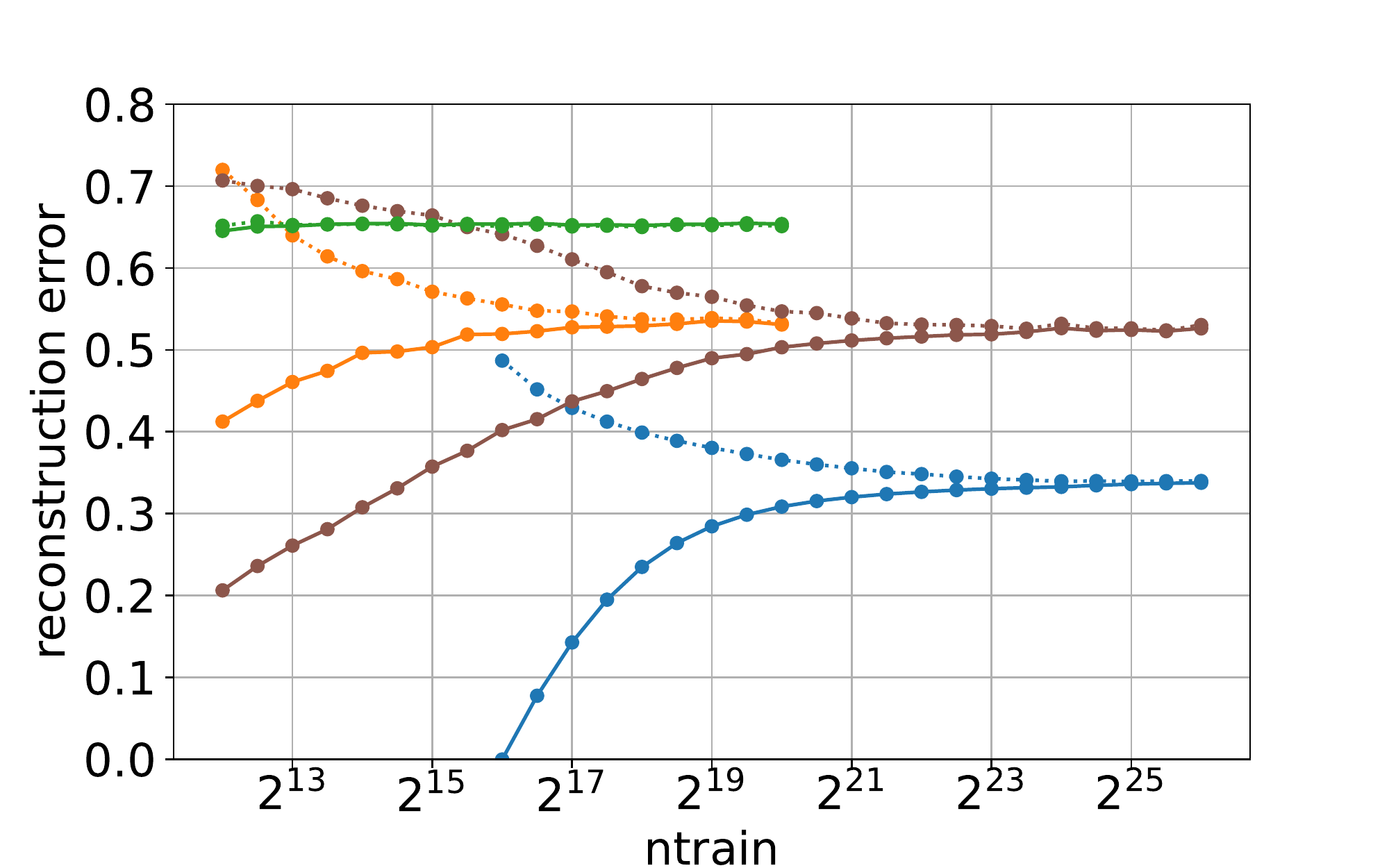}
\hfill
\includegraphics[height=0.27\linewidth,trim=70pt 0 50pt 10, clip]{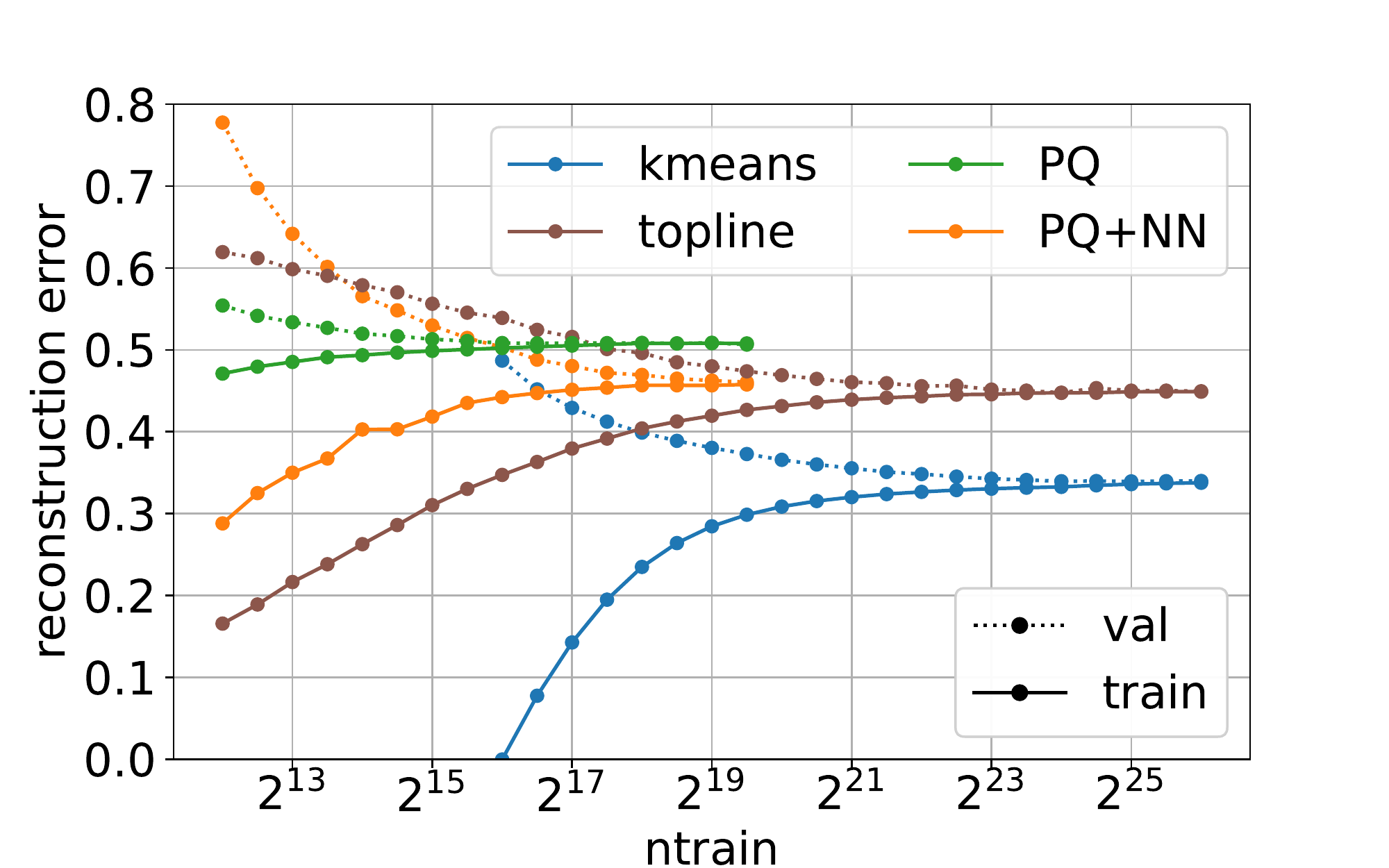}
\hspace{1cm} 
\caption{Small-scale experiment on the Deep1M dataset: reconstruction error when varying the size of the training set with 16-bit codes. We compare the \textcolor{MidnightBlue}{\bf kmeans} baseline with different reconstruction strategies for decoding PQ codes produced by the same PQ encoder: the regular \textcolor{ForestGreen}{\bf PQ} decoding, the \textcolor{Brown}{\bf topline} for a PQ encoding, and our  \textcolor{orange}{\bf PQ+NN} decoder based on a neural network. We  report the MSE on the training set (solid) and on the validation set (dashed). For PQ and PQ+NN, we vary the number $m$ and bits per subspace to keep 16-bit codes: ({\it left}) PQ4$\times$4: $m$\,=\,4, nbits\,=\,4 and ({\it right}) PQ2$\times$8. 
\label{fig:main_deep}}
\end{figure*}

\subsection{AQ: a better decoder for PQ/OPQ} 
Our first proposal is to adopt the Additive Quantization (AQ) decoder of Eqn.~\ref{equ:aq} for PQ, and optimized PQ (OPQ). OPQ is a variant of PQ where, similar to ITQ, the method applies a learned rotation before the subspace partitioning~\cite{ge2013optimized,norouzi2013cartesian}.  The OPQ decoder is identical to PQ except that it rotates the vector back to compensate for the initial rotation.  %
PQ and OPQ are special cases of AQ. Adopting an AQ decoder instead of the usual PQ decoder implies that we consider specific reconstruction LUTs $\Qset'_i$ that have $d$ dimensions instead of ${d}/{m}$: the reconstruction is a summation with Eqn.~\ref{equ:additive_quantizer} instead of a concatenation. Therefore and in contrast to existing quantization-based methods, we disentangle the look-up tables associated with the encoder from the ones associated with the decoder: we have two sets of look-up tables. 

This alleviates the decoding constraint of PQ, where each subindex only contributes to the reconstruction in its own subspace. Since the subspace are not totally independent, even after application of a pre-rotation like OPQ, the AQ decoder improves the reconstruction.

\subsection{Binary codes: LUTs reconstruction} 
We also propose to adopt the AQ decoder of Eqn.~\ref{equ:aq} to reconstruct binary codes. 
While the decoding procedure is conceptually identical to the case of PQ and OPQ, in this binary context this choice departs significantly from the current practice in the literature.  where there is usually no reconstruction procedure associated with the binarization, or only a simplistic one. 

In our case, for a $m$-dimensional bit vector, we learn $m$ LUTs of size $d \times 2$. Each LUT is indexed by a bit value $k_i$ as $\Qset_i[k_i]$. 
To our knowledge it is the first time that the AQ (strong) decoder is proposed for binarization techniques. 
It is advantageously combined with ADC to avoid any approximation on the query. As we will see, it provides a significant improvement without extra memory and at a negligible compute-cost when used for re-ranking. 
The only requirement compared to usual binary codes is that the comparison is not context-free: we need to store the lookup tables  $\Qset_i$ to enable the comparison between a query and a vector, in contrast to the context-free Hamming distance comparison. 

This stronger decoder for binary code is backward-compatible in the sens that it can be applied for an existing index of binary codes, with the following requirement: one needs a training set of vectors and corresponding binary codes, which are required to learn the LUTs with Eqn~\ref{equ:aq}.  

%

%
%

\subsection{Discussion}  

AQ is the best possible decoder with linear reconstruction as in Eqn.~\ref{equ:additive_quantizer}. 
In the literature, different AQ methods differ by how the encoding is performed, which impacts the trade-off between speed and encoding time. 
However, those offering the best trade-offs like LSQ++ are computationally intensive. In Figure~\ref{fig:encoding_time} we plot the compromise between encoding time %
and mean squared error (MSE). 

\mypar{LSQ++ vs PQ8$\times$8.}  
The LSQ++ encoder ($m=8, K'=256$) is 2 orders of magnitude slower than its PQ8$\times$8 counterpart. It is also significantly better than PQ. However, with our PQ+AQ, that combines an AQ decoder with a PQ encoder, the gap is reduced significantly. This advocates the choice of a faster encoder. 

\mypar{PQ8$\times$8 vs PQ16$\times$4.} 
The relatively poor reconstruction accuracy associated with a PQ16$\times$4  decoder, when using the corresponding naive decoder, is significantly improved with AQ decoding: it even outperforms PQ8$\times$8 while being one order of magnitude faster, due to the much lower number of centroids per subquantizers (16 versus 256). A key advantage of PQ16$\times$4 is a strong architectural advantage at search time: The look-up table $\Qset_i$ can be stored in the process registries~\cite{andre2019quicker}, leading to an even larger gap in efficiency. Our proposal to leverage such efficient implementation makes this parameter an appealing choice. 

\subsection{Neural Network decoder} 
\label{sec:nn_decoder} 

The AQ decoder significantly improves binary codes or product quantization encoders. 
However the reconstruction linearly depends  on the separate reconstructions of the components $\Qset_i$. 
This is suboptimal: for instance, binary and PQ reconstruct each sub-vector independently of the others, implicitly assuming independence of the codes ${\mathbb P}(k)=\prod_{i=1}^m {\mathbb P}(k_i)$.
This independence would be true if the encoding was optimal (there would be no redundant information between the $m$ sub-vectors), but is not true in practice (sub-vectors are not independent).
We address this problem by defining a neural network decoder $g$ that, given a compound index $(k_1,\dots,k_m)$, produces a reconstruction from the index, as shown in Figure~\ref{fig:nn_linear}. 
The first layer is a structured LUT similar to $\Qset$ for which we adopt the same notation as PQ:  LUT$m\times b$ indicates that  the tensor implementing this layer contains $m  \times 2^b \times d$ weights. 
In our experiments, we use LUT16$\times$4 and LUT8$\times$8 with PQ16$\times$4 and PQ8$\times$8, respectively. For 64-bit binary codes we use LUT64$\times$1. 

After the first layer of the decoder (LUT parameters), we stack one or more blocks. 
Each block consists of a batch normalization and two fully connected layers separated by a ReLU activation function. 
In the following, we restrict this network to one block, as we observed empirically that more blocks did not provide significant improvements.
Note that if the decoder consisted of one LUT followed by an addition, it would be equivalent to the AQ decoder. 
\subsection{Triplet loss} 
\label{sec:tripletloss}

We optionally consider the triplet loss as an additional term to preserve more explicitly the initial ranking in the reconstruction space, defined as%
\begin{equation}
\mathcal{L}_{\textrm{triplet}} = \max(0, \|x-q(x^{+})\|^2_2 - \|x-q(x^{-})\|^2_2 + \delta). 
\end{equation}
In this equation, we consider a query $x$, a positive match $x^{+}$ in a given neighborhood (defined by rank)  and a negative match $x^{-}$ selected to be a hard negative. 
The margin $\delta$ ensures separation between positives and negatives and prevents the weights from collapsing to zero.
The overall loss combines the triplet loss and the reconstruction loss, as:

\begin{equation}
\mathcal{L} = \mathcal{L}_{\textrm{recons}} + \lambda \cdot \mathcal{L}_{\textrm{triplet}}
\end{equation}
where $\mathcal{L}_{\textrm{recons}} = \|x-q(x) \|^2$ is the reconstruction loss and
the parameter $\lambda \geq 0$ controls the trade-off between reconstruction and ranking quality. 
We vary the parameter $\lambda$ to identify the optimal values where we reach the best recall scores. 
We retain the range of $\lambda$ values for which we get the best 100 recall@1. 
For our two test datasets, a value of $\lambda=1$ gives near-optimal results. 

\section{Analysis: A preliminary experiment} 
\label{sec:preliminary}

While the objective of this paper is to improve the performance of indexing techniques based on compact codes, we first evaluate our proposal to change the decoder on a vanilla quantization task.

The encoder $f$ is the stage that defines the space partitioning.  
For a given encoder, the optimal decoder $g$ is known and given by Eqn.~\ref{equ:optimal_reconstruction}: we refer to it as the ``topline''. It can be implemented as a lookup table containing the $K$ $d-$dimensional centroids.
 In practical settings ($K>2^{32}$) the topline computation is not feasible.
To circumvent this limitation, we consider a scale where it \emph{is} feasible to estimate the optimal quantizer: we set $K=2^{16}$ for the total number of centroids, \ie we consider 16-bit codes.

\subsection{Setup of the experiment} 

At that scale it is possible to run the full k-means quantizer. 
Therefore we can compare the following encoders: 
\begin{itemize}
\item  
    the k-means encoder that groups data points in $k=2^{16}$ clusters.
    This is the topline encoder for the training set because it minimizes the MSE itself;
\item 
    PQ2$\times$8 splits vectors into 2 sub-vectors, each encoded in 8 bits. 
    This is a constrained setting of the k-means encoder, because it is less general; 
\item 
    PQ4$\times$4 splits vectors into 4 sub-vectors, each encoded in 4 bits. 
    This setting is even more constrained. 
\end{itemize}

For PQ encoders, we compare the decoders: 
\begin{itemize}
\item 
    the ``natural'' decoder uses the PQ tables to reconstruct the vectors, \emph{i.e.,} those used by the PQ encoder;
\item 
    the topline decoder uses a $K\times d$ size lookup table with the optimal reconstruction from Equation~(\ref{equ:optimal_reconstruction}). 
    Note that this setting is feasible only in a very small scale like here;
\item 
    the neural net (NN) decoder reconstructs the vectors with a small neural net, see Section~\ref{sec:nn_decoder}.
\end{itemize}

The k-means encoder can be seen as the product quantizer PQ1$\times$16. 
For PQ1$\times$16 the ``natural'' and the topline decoders coincide. 
In addition, the NN decoder of a PQ1$\times$16 also boils down to a look-up table because all possible inputs of the NN are mapped into a table. 

In Figure~\ref{fig:main_deep}
we measure the MSE 
as a function of the number of training vectors. Note that the linear additive decoder (PQ
as encoder and AQ as decoder) was omitted because AQ is a particular case of the neural network decoder: the AQ decoder is equivalent to our NN decoder with just the LUT layer.

\subsection{Training and validation error}

In all the  settings, when increasing the number of training vectors, we observe the typical behavior of learning algorithms: 
for few training vectors, the MSE on training is much lower than that on validation vectors (overfitting); for more training vectors, the two errors become identical.
This is because of the generalization capacity to unseen data of any algorithm trained on more data.

This transition from overfitting to convergence  occurs for all encoder/decoder pairs, but the speed of convergence depends on the capacity of the encoder and decoder: for ``natural'' decoders it is faster for PQ2x8 than for k-means because the latter has more parameters to train. 
It is even faster for PQ4$\times$4. 
On the decoder front, the topline decoder has the same number of parameters as the regular k-means, so it is not suprising that both converge as slowly. 
The NN decoder is in-between the topline and natural decoders.

\subsection{Discussion}

%

 The linear additive decoder achieves at best the same performance as the optimized decoder (PQ as encoder and NN as decoder).

We first compare the ``topline'' curves with the k-means curves. 
This quantifies the 
suboptimality of the encoder because the k-means is an optimal encoder and decoder while the topline has a PQ encoder with an optimal decoder. 
The difference with the topline is much higher for PQ4$\times$4, which is a particular case (and more constrained) of PQ2$\times$8. 
Then we compare the ``topline'' with the ``natural'' PQ decoder. 
This shows the contribution of the decoder only. 
We observe that the gain due to the encoder is a bit smaller than that due to the decoder. 

By adding an optimized decoder after the encoding step, we attempt to approach the optimal decoder with a NN that scales beyond this toyish setup. 
We observe that the NN decoder has an asymptotic accuracy close to that of the topline decoder.

Interestingly our PQ+NN decoder, while asymptotically (ntrain$\rightarrow \infty$) inferior to the topline, achieves better performance on the validation set than the topline in the data-starving regime. Our interpretation is that it has to learn fewer parameters and is therefore better able to generalize with less data.

\label{arch-nn}

%% file: experiments.tex
\section{Experiments}
\label{sec:experiments}

\begin{figure}
    ~\hfill
    \includegraphics[width=1\linewidth,trim=0 20 0 20]{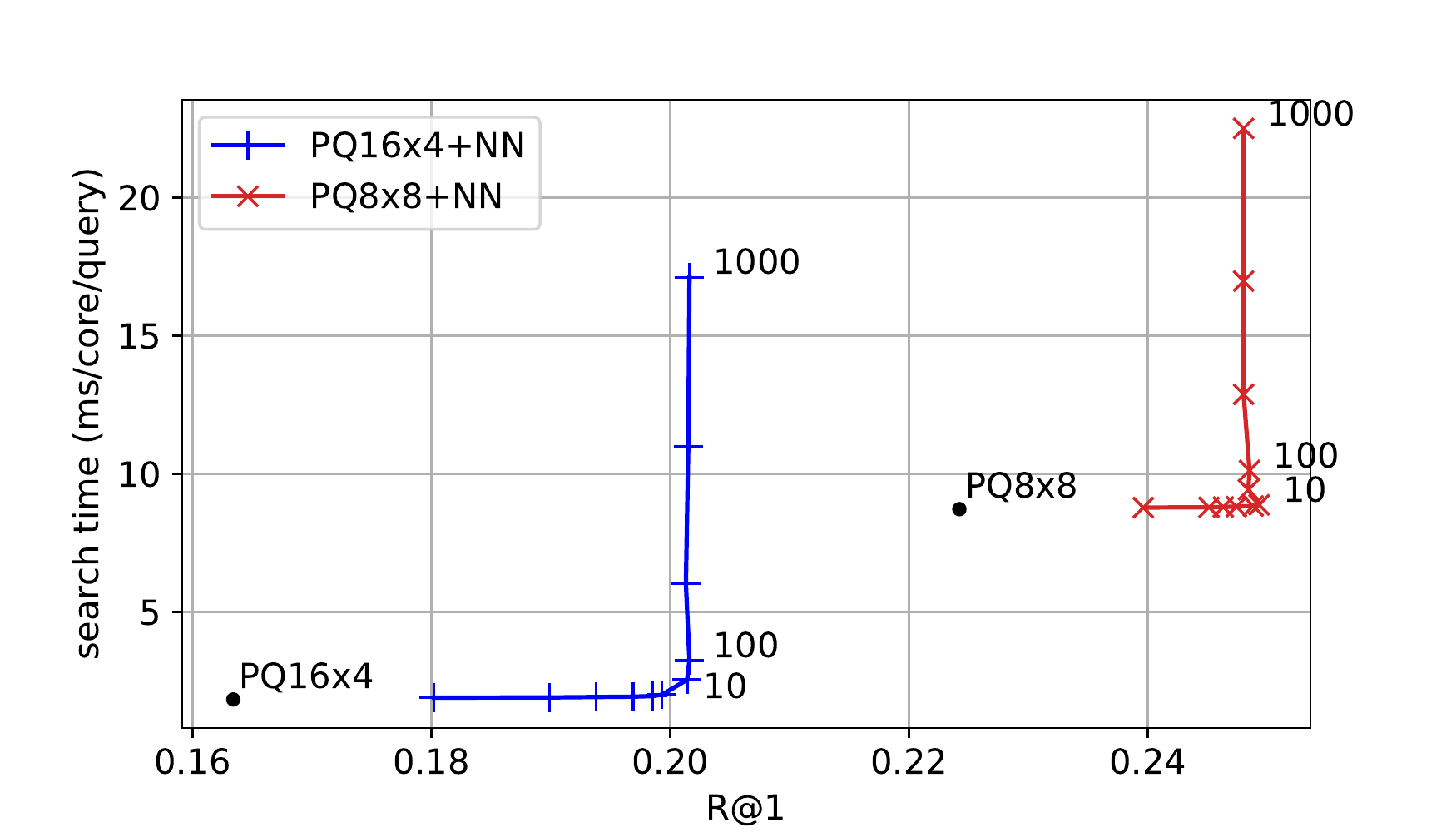}
    \hfill ~
    \caption{
    Accuracy vs. search time on the BigANN1M dataset when re-ranking:
   The NN decoder re-orders the top PQ results.
    The isolated points correspond to the baseline, i.e., without reranking the short-list.
    The curves are obtained by sweeping over the number of top elements to re-rank (2 to 1000).
    %
    \label{fig:reranking}
    }
\end{figure}

\subsection{Experimental setting} 
\label{sec:experiment_setting}
We use publicly available benchmarks to evaluate the performance of nearest neighbor search techniques, namely \textbf{BigANN1M}~\cite{jegou2011searching} ($d=128$) and \textbf{DEEP1M}~\cite{babenko2016efficient} ($d=96$). 
Both are image features extracted from real images, arranged in a database of 1M vectors, a query set of 10.000 queries, and a separate set of training vectors. 
We measure the Recall@$R$, i.e. the rate of queries for which the nearest neighbor is ranked in the first $R$ ranks, for a code of size 64 bits in all the experiments. 
The measurements are averaged over 5 runs of training with different random seeds. 
Our NN decoder minimizes the reconstruction loss with Adam optimizer. 
We train on 300 epochs with a batch size $\textrm{bs}=256$ and a learning rate $\textrm{lr} = 5\cdot 10^{-4}$. 
We use a scheduler that reduces the learning rate by a factor $\textrm{lr}_{\textrm{decay}} = 0.5$ when the validation loss stops improving. %
We do not regularize with weight decay.

\subsection{Results with PQ codes}
Table~\ref{tab:pqresults} compares the deep decoder with baselines in terms of recall for PQ/OPQ encodings. 
The AQ decoder already improves the accuracy with respect to the PQ/OPQ baseline. 
Wwe obtain the largest improvement with the neural network (NN) decoder, especially for PQ16$\times$4 codes. This parameter choice seems of high practical interest, since it combines a very fast encoder with a competitive indexing performance.

\mypar{Re-ranking.} 
We use this approach in a re-ranking setting: 
since we have a fast decoder (row with Decoder ``PQ") and a slower but more accurate one (Decoder ``NN"), we consider a two-stage retrieval procedure, where we first filter out at least 99.9\% of the vectors with the fast one. 
 
Figure~\ref{fig:reranking} shows the results of this approach. 
Most of the accuracy gain is obtained by re-ranking just the top-10 first-level results. Therefore the re-ranking time is negligible w.r.t. the initial search time. 
The largest gain (3.4 points) is obtained with PQ16$\times$4 codes, that are also the fastest for the first-level decoder.

\def \mysp {\ \ \ \ }
\begin{table*}[t]
\caption{\label{tab:pqresults}
Retrieval results on BigANN1M and Deep1M with PQ/OPQ 64-bit quantization and ADC comparisons. 
The LSQ++ results are a topline, very slow at encoding time. 
All results are computed with ADC. 
The OPQ (en/de)coder is identical to PQ (en/de)coder up to a learned rotation. 
The methods introduced in this paper are\,\colorbox{blue!10}{shaded.} %
}\centering
{
\begin{tabular}{c@{\mysp}c|r@{\mysp}r@{\mysp}r|r@{\mysp}r@{\mysp}r}
\toprule
& & \multicolumn{3}{|c|}{\textbf{16$\times$4}} &      \multicolumn{3}{c}{\textbf{8$\times$8}}         \\ 
\cmidrule(lr){3-5} \cmidrule(lr){6-8}
        Encoder & Decoder & R@1 & R@10 & R@100  & R@1 & R@10 & R@100 \\ 
\toprule
\multicolumn{8}{c}{\textbf{BigANN1M}}                                                     \\ 
\midrule
PQ & PQ                     & 0.168            & 0.530            & 0.887            & 0.223             & 0.651              & 0.948               \\ 
\rowcolor{blue!10} 
PQ & AQ                     & 0.182            & 0.564            & 0.908            & 0.234             & 0.667              & 0.955               \\ 
\rowcolor{blue!10} 
PQ & NN                     & 0.202            & 0.606            & 0.928            & \textbf{0.239}    & \textbf{0.681}     & 0.958      \\ 
OPQ & OPQ                   & 0.194            & 0.605            & 0.937            & 0.231             & 0.667              & \textbf{0.960}               \\ 
\rowcolor{blue!10} 
OPQ & NN                    & \textbf{0.202}   & \textbf{0.621}   & \textbf{0.945}   & 0.225             & 0.665              & 0.959               \\ \midrule
LSQ++ & AQ                  &                  &                  &                  & \textbf{0.309}    & \textbf{0.785}     & \textbf{0.987}      \\ 
\toprule
\multicolumn{8}{c}{\textbf{Deep1M}}                                                       \\ 
\midrule
PQ & PQ                     & 0.087           & 0.324             & 0.703            & 0.091             & 0.339              & 0.730                \\ 
\rowcolor{blue!10} 
PQ & AQ                     & 0.083           & 0.313             & 0.670            & 0.094             & 0.355              & 0.749                \\ 
\rowcolor{blue!10} 
PQ & NN                     & 0.100           & 0.370             & 0.756            & 0.105             & 0.380              & 0.776                \\ 
OPQ & OPQ                   & 0.151           & 0.493             & 0.872            & 0.167             & 0.538              & 0.898                \\ 
\rowcolor{blue!10} 
OPQ & NN                    & \textbf{0.154}  & \textbf{0.516}    & \textbf{0.889}   & \textbf{0.168}    & \textbf{0.550}     & \textbf{0.908}       \\ \midrule
LSQ++ & AQ                  &                 &                   &                  & 0.246   & 0.688     & 0.965       \\ 
\bottomrule
\end{tabular}}
\end{table*}

\subsection{Results on binary codes}
Table~\ref{tab:exp_binary} reports results with binary encoders. 
We consider two encoders: ITQ~\cite{gong2012iterative} and the catalyzer~\cite{sablayrolles2018spreading}. 
We show how the deep decoder stands amongst popular baselines in term of reconstruction error and recall for binary encodings. Recall that for the AQ solver and the optimized decoder, the lookup table structure is M$\times$$b$ = 64$\times$1. 
With binary codes, an asymmetric comparison is the element that provides the most significant boost in accuracy, which is shown by the comparison between SDC and ADC. 

Our approach ITQ+NN provides an additional gain compared with the ITQ encoder. For the stronger encoder (catalyzer), our approach catalyzer+NN provides a significant improvement on the Deep1M dataset, in particular when adding a triplet loss to make our training more consistent with the one of the catalyzer. However we point that on BigANN1M, our simpler choice of using AQ as a decoder is the best. This may be due to the optimization because formally, the AQ decoder is a particular case of the NN decoder.

\begin{table*}
\caption{Performance of different binary quantizers (64\,bits), on Deep1M and BigANN1M, with ITQ encoding or the neural network encoder of~\cite{sablayrolles2018spreading}. 
SDC refers to the case where we compare codes with Hamming distance,  ADC when the database vector is reconstructed by Eqn.~\ref{equ:recons_binary}. AQ and NN decoders also use an asymmetrical comparison. The row NN/triplet corresponds to the case where we combine the $\ell_2$ loss with a triplet loss similar to the one used to learn the catalyzer, as discussed in Section~\ref{sec:tripletloss}. 
The methods introduced in this paper are\,\colorbox{blue!10}{shaded,} see Table~\ref{tab:alternative_Decoders}. 
\label{tab:exp_binary}
}
\centering
\scalebox{1.}{
\begin{tabular}{cc|rrr|rrr}
\toprule
 &  &  \multicolumn{3}{c}{\textbf{BigANN1M}} & \multicolumn{3}{c}{\textbf{Deep1M}}       \\
\midrule
Encoder   & Decoder&  R@1  & R@10   & R@100  & R@1     & R@10   & R@100  \\ 
\midrule
ITQ       & SDC  & 0.055          & 0.220          & 0.538           & 0.056          & 0.213          & 0.516          \\ 
ITQ       & ADC  & 0.103          & 0.383          & 0.783           & 0.100          & 0.368          & 0.759          \\ 
\rowcolor{blue!10} 
ITQ       & AQ   & 0.098          & 0.372          & 0.768           & 0.097          & 0.362          & 0.753          \\ 
\rowcolor{blue!10} 
ITQ       & NN   & \textbf{0.118} & \textbf{0.427} & \textbf{0.819}  & \textbf{0.112} & \textbf{0.401} & \textbf{0.790} \\ 
\midrule 
catalyzer & SDC  &      0.083 &	0.298 &	0.622                       &  0.071         & 0.254          & 0.558  \\ 
catalyzer & ADC  & 0.158          & 0.520          & 0.879           &  0.137         & 0.457          & 0.830  \\ 
\rowcolor{blue!10} 
catalyzer & AQ   & \textbf{0.160} & \textbf{0.524} & \textbf{0.881}  &  0.139         & 0.459          & 0.833  \\ 
\rowcolor{blue!10} 
catalyzer & NN   & 0.153          & 0.509          & 0.865           &  0.142         & 0.463          & 0.834  \\ 
\rowcolor{blue!10} 
catalyzer & NN/triplet    & 0.157 & 0.519           & 0.876          & \textbf{0.145} &\textbf{ 0.471} & \textbf{0.841}  \\ %
\bottomrule
\end{tabular}
}
\end{table*}

\subsection{Other limiting factors}

For most applications there is a single limiting factor.
In this paper we mainly fix the code size and evaluate the encoding accuracy vs. speed tradeoff. 
However, there are other resource constraints that can become limiting. 
Concerning the memory requirement of storing the codes of the look-up tables itself, the neural network approach is less parsimonious than fixed (binary) quantizers that don't need to store centroids in lookup tables. 
The parameters of a neural network decoder exceed that of a linear additive quantizer because they store LUTs, and also the trained network parameters. 
Note that the memory usage for the codec is rarely a limiting factor because it is constant w.r.t. the amount of data to process. 

The optimized decoder added to a fixed PQ encoder is always the optimal solution in term of accuracy given a fixed encoding time.
Note that the NN decoder
training time is not a problem in this context: it is several orders of magnitude faster than training
image classification networks and can easily be done on CPU.

\subsection{Sensitivity to decoder parameters}

We analyse the sensitivity of our neural network decoder to variations of the hyper-parameters. 
We run two analyses: one on the network architecture and the other on the parameters in the decoder training process. 
In our analysis, all results have been run on the BigANN1M dataset, with a training set of $500000$ points and a validation set of $100000$ points. 
The inputs of the network are codes returned by a PQ16$\times$4 encoder.

\begin{figure}
\centering

 \begin{subfigure}[b]{0.55\textwidth}
       \includegraphics[width=0.89\linewidth,trim=0 0 0 0, clip]{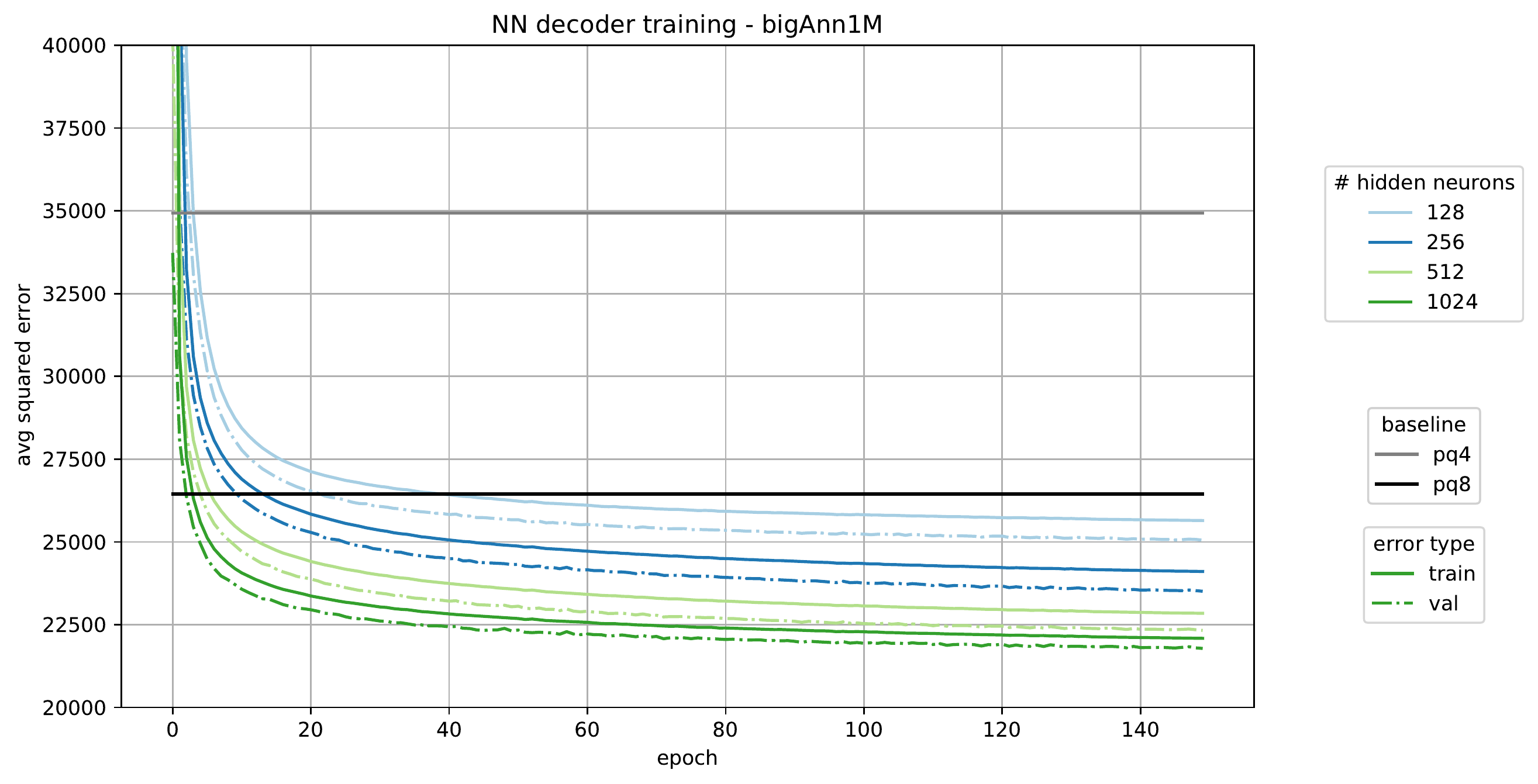}
       \subcaption{\centering Number of neurons in hidden layers}
       \label{fig:gridsearch_nblocks}
    \end{subfigure}

~
    
    \begin{subfigure}[b]{0.55\textwidth}
       \includegraphics[width=0.89\linewidth,trim=0 0 0 0, clip]{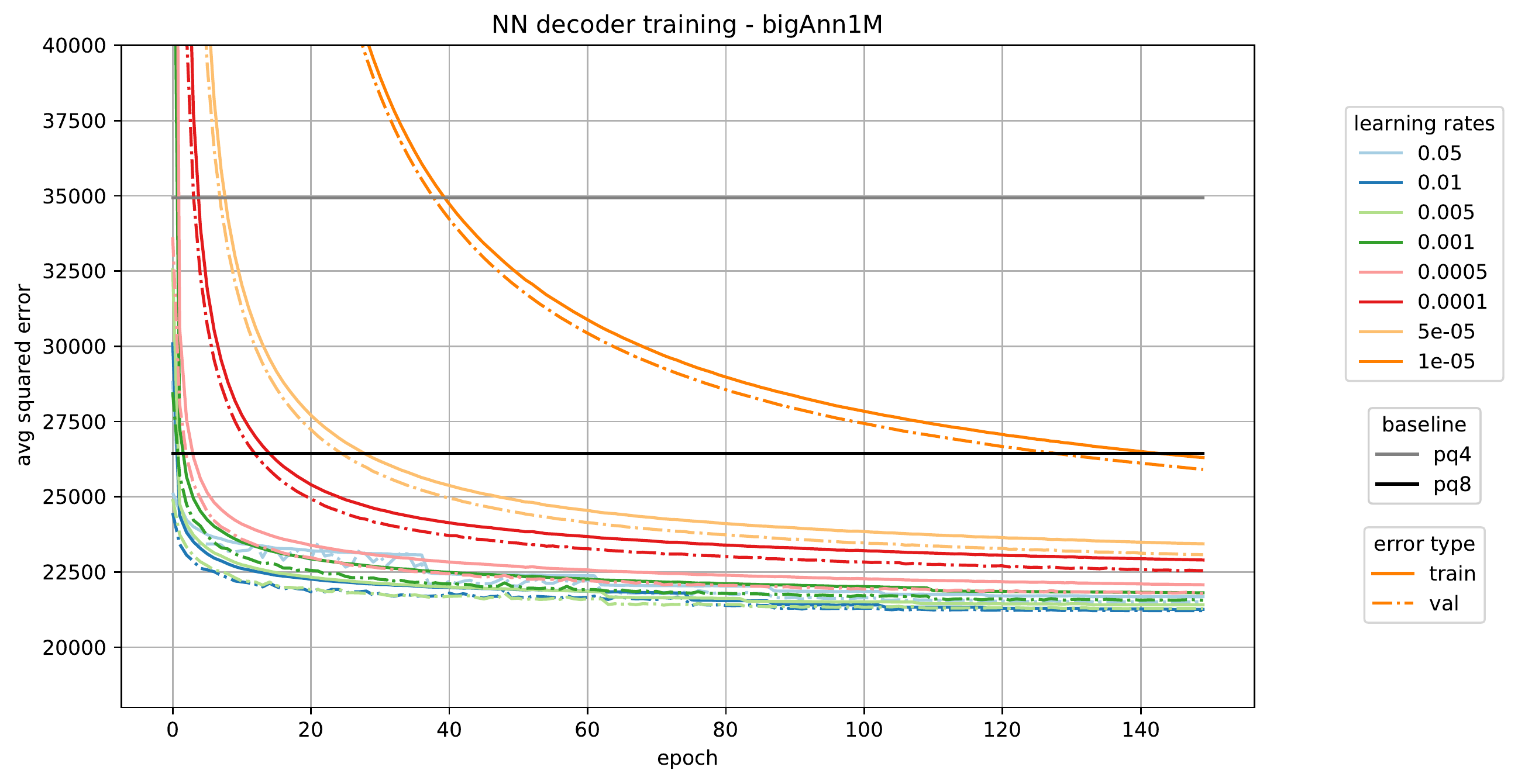}
       \subcaption{\centering Learning rates}
       \label{fig:gridsearch_lr} 
    \end{subfigure}

~    

    \caption{Training and validation losses during training of the optimized decoder for different hyperparameters: (a) number of neurons in hidden layers and (b) learning rates. The decoder is a linear neural network with 2 blocks, trained with Adam optimizer on the BigANN1M dataset, with a training set of $500$k points and a validation set of $100$k points. }
    \label{fig:gridsearch}
\end{figure}

\mypar{Architecture.}
The parameters we consider for the network architecture are the type of blocks (linear or residual), the number of blocks, the number of neurons in the hidden layers, and the dropout rate. 
In our case, a residual architecture doesn't significantly improve the performance of the decoder, probably because the depth of the network is low (only 1 to 3 blocks). 
We observe on Figure~\ref{fig:gridsearch_nblocks} that a linear network with only 2 blocks already outperforms AQ decoders and adding more blocks shows down the inference without adding much more accuracy.

Whatever the number of neurons in the hidden layers, the learning process of the decoder is stable but we reach smaller loss values with more hidden neurons. 
We experimented with dropout but it did not improve the validation accuracy significantly.

\mypar{Optimization.}
The parameters we consider for the decoder optimization are  the optimizer, the learning rate, the learning decay factor, the weight decay factor, and the batch size. 
We compare four optimizers that are commonly used in deep learning: SGD, Adam, Adadelta and RMSprop. 
We vary the learning rate from $5\cdot10^{-3}$ to $5\cdot10^{-5}$ to assess their stability. 
We observe that all networks have a stable learning process and achieve their best accuracy scores for different range of learning rates. 
We choose Adam optimizer because it is more locally stable and was shown to be faster and more stable than SGD when fine-tuned~\cite{choi2019empirical}. 

Having chosen Adam as the optimizer, we vary more precisely the learning rates. 
Figure \ref{fig:gridsearch_lr} shows that both training and validation losses smoothly decrease and that the decoder is stable with regards to variations of the learning rate.
We recommend to use learning rates greater than $5\cdot10^{-4}$. They reach a better accuracy than the fixed encoder/decoder PQ in less than 5 epochs. 
We tested the effect of learning rate decay: varying the learning rate decay factor from 0.2 to 1 has no significant effect on the learning process.
We draw similar conclusions when varying the weight decay factor from 0 to 0.2. The batch size has no significant influence on the decoder training. After epoch 40, the optimization reaches the same loss values whatever the batch size (128, 256, 512 and 1024).

Overall, since the optimization does not appear to be sensitive to hyperparameters, we select the most lightweight architecture and the most natural hyperparameters for all our experiments (see Section~\ref{sec:experiment_setting}).